\documentclass{article}

\PassOptionsToPackage{numbers,compress}{natbib}

\usepackage[preprint]{neurips_2025}

\usepackage[utf8]{inputenc} 
\usepackage[T1]{fontenc}    
\usepackage{hyperref}       
\usepackage{url}            
\usepackage{booktabs}       
\usepackage{amsfonts}       
\usepackage{nicefrac}       
\usepackage{microtype}      

\usepackage[utf8]{inputenc} 
\usepackage[T1]{fontenc}    
\usepackage{hyperref}       
\usepackage{url}            
\usepackage{booktabs}       
\usepackage{amsfonts}       
\usepackage{nicefrac}       
\usepackage{microtype}      
\usepackage{xcolor}         
\usepackage{times}
\usepackage{latexsym}
\usepackage{amsmath}
\usepackage{bm}
\usepackage{bbm}
\usepackage{multicol}
\usepackage{graphicx}
\usepackage{subfigure}
\usepackage{multirow}
\usepackage{wrapfig}
\usepackage{algorithm}
\usepackage[noend]{algpseudocode}

\usepackage{multirow}
\usepackage{array}
\newcommand{\PreserveBackslash}[1]{\let\temp=\\#1\let\\=\temp}
\newcolumntype{C}[1]{>{\PreserveBackslash\centering}p{#1}}
\newcolumntype{R}[1]{>{\PreserveBackslash\raggedleft}p{#1}}
\newcolumntype{L}[1]{>{\PreserveBackslash\raggedright}p{#1}}
\usepackage{diagbox}
\usepackage{booktabs}
\usepackage{graphicx}
\usepackage{wrapfig}
\usepackage{caption}
\usepackage{subcaption}
\usepackage{multirow}
\usepackage{bbm}
\usepackage{enumitem}
\usepackage{amsmath,amsfonts,bm}
\usepackage{tcolorbox}

\title{Stream-Omni: Simultaneous Multimodal Interactions with Large Language-Vision-Speech Model}

\author{
Shaolei Zhang\textsuperscript{\rm 1,3},\quad Shoutao Guo\textsuperscript{\rm 1,3},\quad Qingkai Fang\textsuperscript{\rm 1,3},\quad Yan Zhou\textsuperscript{\rm 1,3},\quad Yang Feng\textsuperscript{\rm 1,2,3}\thanks{Corresponding author: Yang Feng.} \\
\textsuperscript{\rm 1}{Key Laboratory of Intelligent Information Processing,} \\ \textsuperscript{$\;\;$}Institute of Computing Technology, Chinese Academy of Sciences (ICT/CAS) \\
{\textsuperscript{\rm 2}{Key Laboratory of AI Safety, Chinese Academy of Sciences}} \\
{\textsuperscript{\rm 3}{University of Chinese Academy of Sciences, Beijing, China}} \\
\textsuperscript{$\;\;$}\texttt{\href{mailto:zhangshaolei20z@ict.ac.cn}{zhangshaolei20z@ict.ac.cn}, \href{mailto:fengyang@ict.ac.cn}{fengyang@ict.ac.cn}}}

\begin{document}

\maketitle

\vspace{-3mm}
\begin{abstract}
\setcounter{footnote}{0}
The emergence of GPT-4o-like large multimodal models (LMMs) has raised the exploration of integrating text, vision, and speech modalities to support more flexible multimodal interaction. Existing LMMs typically concatenate representation of modalities along the sequence dimension and feed them into a large language model (LLM) backbone. While sequence-dimension concatenation is straightforward for modality integration, it often relies heavily on large-scale data to learn modality alignments. In this paper, we aim to model the relationships between modalities more purposefully, thereby achieving more efficient and flexible modality alignments. To this end, we propose Stream-Omni, a large language-vision-speech model with efficient modality alignments, which can simultaneously support interactions under various modality combinations. Stream-Omni employs LLM as the backbone and aligns the vision and speech to the text based on their relationships. For vision that is semantically complementary to text, Stream-Omni uses sequence-dimension concatenation to achieve vision-text alignment. For speech that is semantically consistent with text, Stream-Omni introduces a CTC-based layer-dimension mapping to achieve speech-text alignment. In this way, Stream-Omni can achieve modality alignments with less data (especially speech), enabling the transfer of text capabilities to other modalities. Experiments on various benchmarks demonstrate that Stream-Omni achieves strong performance on visual understanding, speech interaction, and vision-grounded speech interaction tasks. Owing to the layer-dimensional mapping, Stream-Omni can simultaneously provide intermediate text outputs (such as ASR transcriptions and model responses) during speech interaction, offering users a comprehensive multimodal experience\footnote{Code: \href{https://github.com/ictnlp/Stream-Omni}{https://github.com/ictnlp/Stream-Omni}, Model: \href{https://huggingface.co/ICTNLP/stream-omni-8b}{https://huggingface.co/ICTNLP/stream-omni-8b}.}.

\end{abstract}

\section{Introduction}
\label{sec:intro}

Large multimodal models (LMMs) such as GPT-4o \citep{gpt-4o} exhibit omni-capabilities across text, vision, and speech modalities, unlocking broad potential across applications. Compared to vision-oriented LMMs \citep{gpt-4v,llava}, omni-modal LMMs can support speech interaction based on visual information. Furthermore, advanced online services like GPT-4o can offer a seamless “see-while-hear” interaction for users by simultaneously providing intermediate text (i.e., transcription of user inputs and model responses) during speech interaction, which highlights the importance of building LMMs that can simultaneously support interactions through various modality combinations.

\begin{figure}[t]
\centering
\subfigure[Sequence-dimension concatenation for modality alignments in previous works]{
\includegraphics[width=0.37\textwidth]{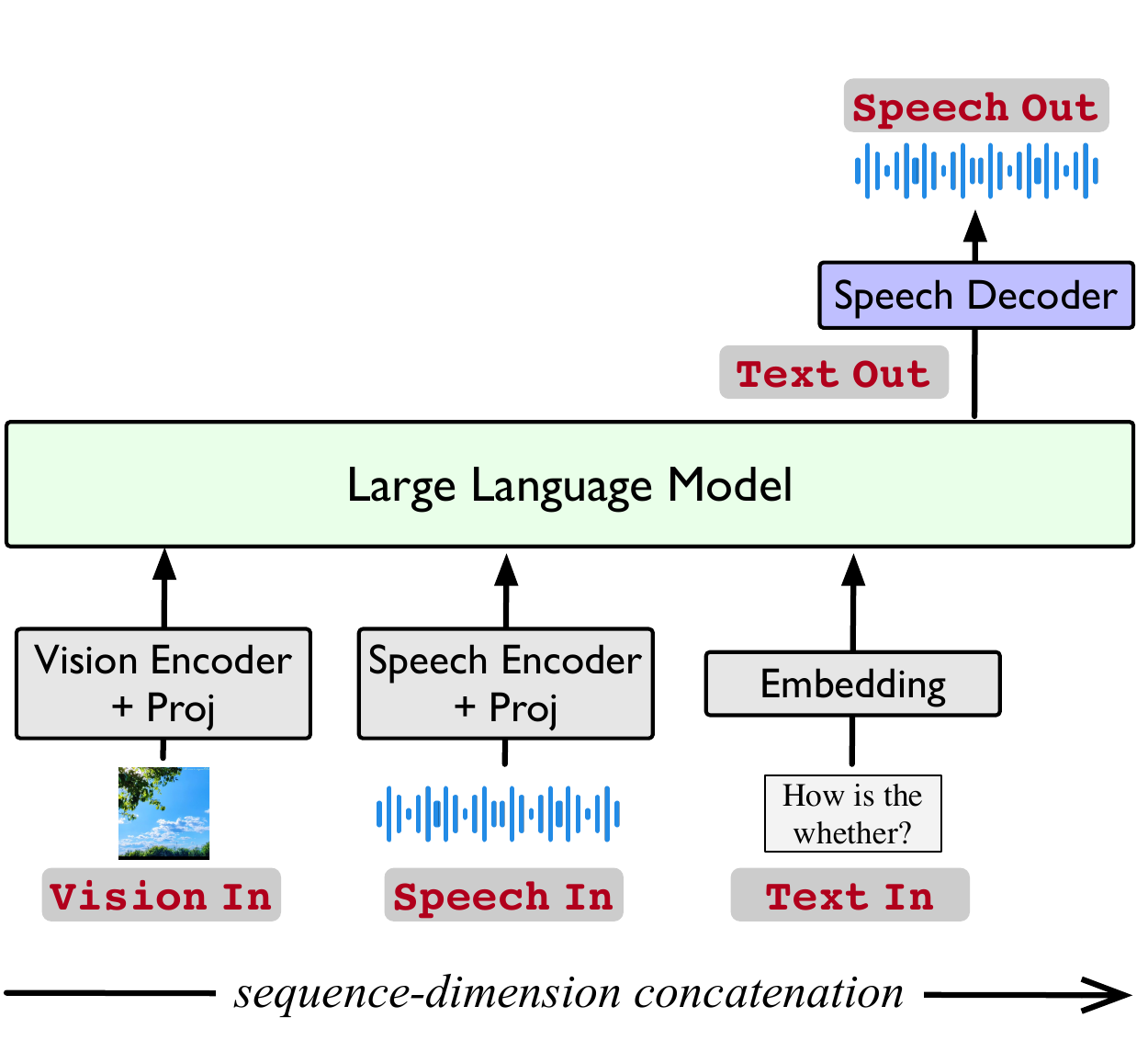} \label{fig:ill1}
}
\quad
\subfigure[Sequence-dimension concatenation for vision-text alignment, and layer-dimension mapping for speech-text alignment]{
\includegraphics[width=0.56\textwidth]{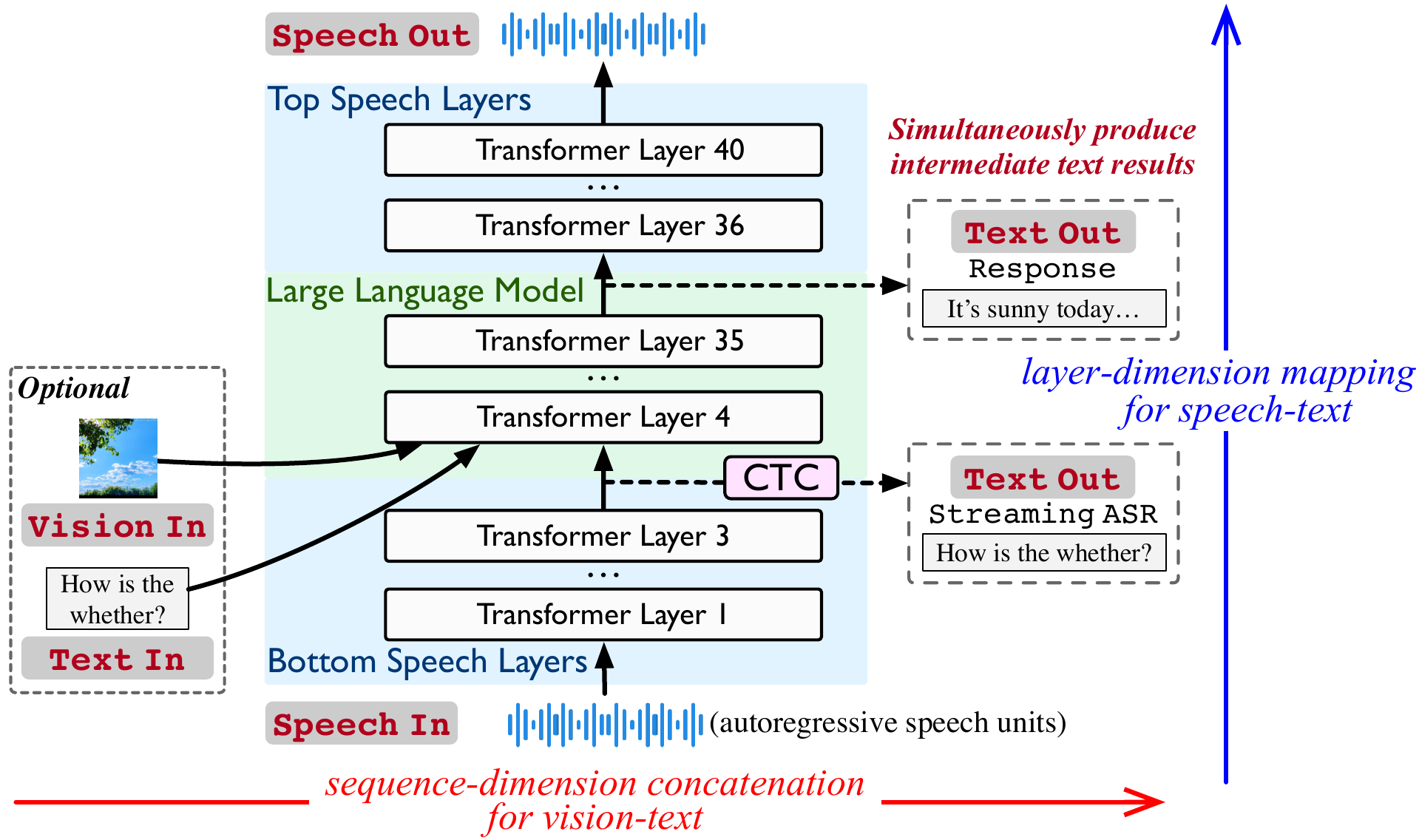} \label{fig:ill2}
}
\caption{Comparison of modality alignments in Stream-Omni and previous works.}
\label{fig:ill}
\end{figure}

However, building LMMs that support text, vision, and speech remains a substantial challenge due to the intrinsic representational discrepancies across modalities.
Most existing LMMs specialize in either vision \citep{minigpt4,llava,liu2024llavanext,li2024llavaonevisioneasyvisualtask,yao2024minicpm} or speech \citep{xie2024miniomnilanguagemodelshear,fang2025llamaomni,défossez2024moshispeechtextfoundationmodel,zeng2024glm4voiceintelligenthumanlikeendtoend}, feeding the extracted modality representations into the context of large language model (LLM) backbone. 
Recently, some omni-modal LMMs \citep{fu2025vita15gpt4olevelrealtime,li2024baichuanomnitechnicalreport,xu2025qwen25omnitechnicalreport} aim to integrate text, vision, and speech within a unified framework. Such models typically concatenate representations from individual modality encoders along the sequence dimension before feeding them into the LLM backbone, as shown in Figure~\ref{fig:ill1}. These concatenation-based approaches simplify modality integration, but they heavily rely on large-scale data to learn modality alignments in a data-driven manner \citep{défossez2024moshispeechtextfoundationmodel,zeng2024glm4voiceintelligenthumanlikeendtoend,li2024baichuanomnitechnicalreport,xu2025qwen25omnitechnicalreport}, which is not friendly to limited public tri-modal data. Moreover, such concatenation-dimension alignments are not flexible enough to simultaneously produce intermediate text results during speech interactions, as GPT-4o does.

To this end, we aim to model the relationships between modalities more purposefully, thereby achieving more efficient and flexible modality alignments. In multimodal interaction, text, vision, and speech modalities serve different roles, where vision primarily conveys visual information \citep{llava}, while text and speech focus on language information \citep{fang2025llamaomni}. As such, directly concatenating all three modalities in sequence-dimension is suboptimal for modality alignments. Ideally, the speech and text should exhibit high semantic consistency, while the vision is semantically complementary to the text. Therefore, vision and speech should be separately aligned to text in different ways.

Along with this idea, we introduce Stream-Omni, a language-vision-speech LMM based on efficient text-centric modality alignments, which can flexibly support interactions under various modality combinations. As shown in Figure~\ref{fig:ill2}, Stream-Omni is built upon the LLM backbone and aligns the vision and speech modalities to text using different mechanisms. For vision, which is semantically complementary to text, Stream-Omni employs sequence-dimension concatenation for vision-text alignment. For speech, which shares higher semantic consistency with text, Stream-Omni introduces a layer-dimension speech-text mapping for speech-text alignment. Specifically, Stream-Omni takes LLM as the core and introduces bottom and top speech layers to model speech-to-text mapping via Connectionist Temporal Classification (CTC) \citep{ctc}, thereby enabling external interaction through the speech modality and simultaneous internal generation via the text modality. With speech–text mapping, Stream-Omni can transfer the text capability of LLM backbone to the speech modality with less speech data. As a byproduct, Stream-Omni can simultaneously produce intermediate text results (i.e., transcription of instruction and response) during speech interaction, offering a more comprehensive multimodal experience. We evaluate Stream-Omni on various benchmarks covering visual understanding, speech interaction, and vision-grounded speech interaction, and the results demonstrate that Stream-Omni achieves strong performance using only 23,000 hours of speech data.

\section{Related Work}

Existing large multimodal models can be categorized into three types: vision-oriented, speech-oriented, and omni-modal. For vision-oriented LMMs, LLaVA \citep{llava} is the most widely adopted architecture. In LLaVA, a vision encoder (CLIP \citep{clip}) is used to extract visual features from visual inputs, which are then concatenated with the text inputs and fed into LLM to generate text responses. Based on LLaVA, the following works improve the vision-oriented LMMs through improved training data~\citep{liu2024llavanext,bai2023qwenvlversatilevisionlanguagemodel,Chen_2024_CVPR}, enhanced image encoding~\citep{lin2023sphinxjointmixingweights,minigpt4,Ye_2024_CVPR}, and extended video understanding~\citep{wang2022internvideogeneralvideofoundation,li2024videochatchatcentricvideounderstanding,videochatgpt,li2023llamavidimageworth2,videollava}.

For speech-oriented LMMs, existing methods rely on either continuous or discrete speech units. Methods based on continuous representations, such as Mini-Omni \citep{xie2024miniomnilanguagemodelshear}, LLaMA-Omni \citep{fang2025llamaomni}, Freeze-Omni \citep{wang2024freezeomnismartlowlatency}, SALMONN-Omni \citep{yu2024salmonnomnicodecfreellmfullduplex}, and SLAM-Omni \citep{chen2024slamomnitimbrecontrollablevoiceinteraction} use a speech encoder (e.g., Whisper \citep{radford2022robust}) to extract speech features, which are then projected into the LLM's embedding space to facilitate speech understanding. These approaches often incorporate a speech decoder to generate speech responses based on LLM's text outputs. Methods based on discrete units, such as SpeechGPT \citep{zhang-etal-2023-speechgpt}, Moshi \citep{défossez2024moshispeechtextfoundationmodel} and GLM-4-Voice \citep{zeng2024glm4voiceintelligenthumanlikeendtoend}, employ a speech tokenizer \citep{hsu2021hubert,zhang2024speechtokenizerunifiedspeechtokenizer,cosyvoice} to convert speech into discrete units, allowing the LLM to directly understand and generate speech units, which are finally synthesized into speech using a unit-based speech decoder \citep{NEURIPS2020_c5d73680,cosyvoice}. Compared to continuous representations, discrete units can be jointly modeled with text in LLM's context, but they often rely on more speech data for speech pre-training \citep{nguyen-etal-2025-spirit,zeng2024glm4voiceintelligenthumanlikeendtoend}.

Existing omni-modal LMMs, such as VITA-1.5 \citep{fu2025vita15gpt4olevelrealtime}, MiniCPM2.6-o \citep{yao2024minicpm}, Baichuan-Omni \citep{li2024baichuanomnitechnicalreport}, Qwen2.5-Omni \citep{xu2025qwen25omnitechnicalreport}, Megrez-Omni \citep{li2025megrezomnitechnicalreport}, M2-Omni \citep{guo2025m2omniadvancingomnimllmcomprehensive}, Capybara-Omni \citep{ji2025capybaraomniefficientparadigmbuilding}, EMOVA \citep{chen2025emovaempoweringlanguagemodels}, OpenOmni \citep{luo2025openomniadvancingopensourceomnimodal} and Omni-Emotion\citep{yang2025omniemotionextendingvideomllm}, use various encoders to extract the modality representations, which are then concatenated and fed into the LLM to facilitate multimodal understanding, and finally a speech decoder is employed to synthesize speech from the generated text. Overall, most existing omni-modal LLMs adopt concatenation-based architectures and rely primarily on data-driven approaches to model modality alignments.


In this work, we focus on the efficiency and flexibility of modality alignments, with the goal of leveraging limited tri-modal data to develop a large language-vision-speech model that can simultaneously support multimodal interactions under various modality combinations. Unlike previous approaches that rely solely on sequence-dimension concatenation for modality alignment, Stream-Omni adopts a more deliberate design by employing sequence-dimension concatenation for vision-text alignment and layer-dimension mapping for speech-text alignment, thereby achieving efficient and flexible modality alignments. As an additional advantage, the proposed layer-dimension mapping enables Stream-Omni to produce intermediate text results during speech interaction, thereby offering users a more comprehensive multimodal experience. 

\begin{figure}[t]
    \centering
    \includegraphics[width=\linewidth]{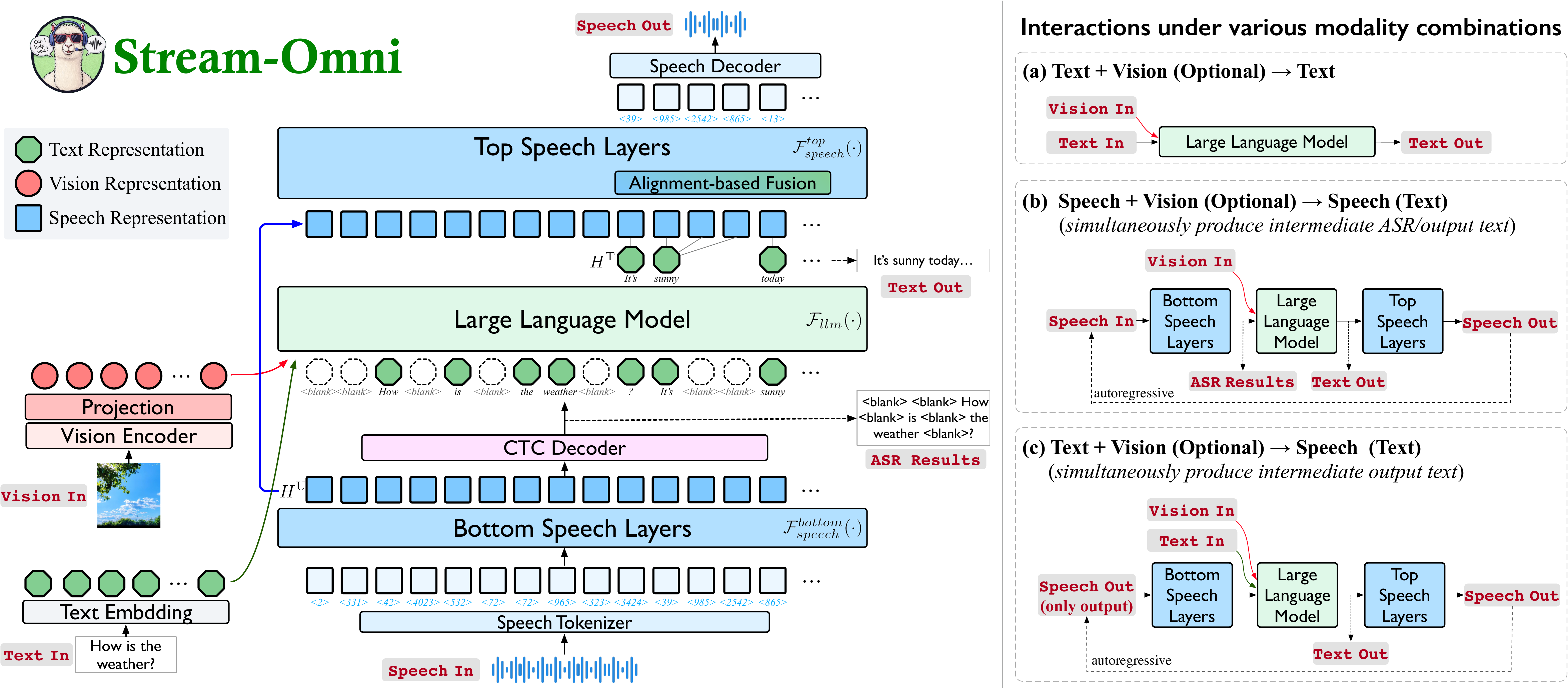}
    \caption{Architecture of Stream-Omni. Right: Interactions under various modality combinations.}
    \label{fig:model}
\end{figure}

\section{Stream-Omni}

We introduce Stream-Omni, a language–vision–speech LMM based on text-centric modality alignments. Stream-Omni aligns vision and speech to the text modality via sequence-dimension concatenation and layer-dimension mapping, respectively, thereby achieving efficient and flexible modality alignments. The architecture, training, and inference of Stream-Omni are introduced as follows.

\subsection{Architecture}

The architecture of Stream-Omni is illustrated in Figure~\ref{fig:model}. Stream-Omni adopts the LLM as its backbone and progressively aligns the vision and speech to the text, efficiently developing a LMM that supports text, vision, and speech. For vision-text alignment, Stream-Omni applies a vision encoder and projection to extract visual representations, which are then concatenated with the text tokens. For speech-text alignment, Stream-Omni introduces several speech layers at the bottom and top of LLM backbone to respectively map speech to the text and generate speech based on the text. 

\subsubsection{Vision Modality}

Given the semantic complementarity between the vision and text modalities, Stream-Omni adopts a sequence-dimension concatenation for vision-text alignment, which is commonly employed in vision-oriented LMMs \citep{llava,liu2024llavanext,llava-mini}. Specifically, Stream-Omni introduces the vision encoder and projection to convert visual inputs into visual representations, which are then concatenated with text representations and jointly fed into the LLM to facilitate visual understanding.

\subsubsection{Speech Modality}

Compared to vision, aligning speech and text is more challenging due to the greater variability of speech representations and the relative scarcity of speech data. To address this, Stream-Omni leverages the higher semantic consistency between speech and text, employing a speech-text mapping to facilitate alignment through more direct supervision.

To achieve this, Stream-Omni incorporates an $N$-layer LLM backbone as the inner core, with $N^{bottom}_{speech}$ speech layers added to the bottom for speech-to-text mapping and $N^{top}_{speech}$ speech layers added to the top for text-to-speech mapping. 
Overall, Stream-Omni extends an $N$-layer LLM into a $(N^{\mathrm{bottom}}_{\mathrm{speech}} + N + N^{\mathrm{top}}_{\mathrm{speech}})$-layer decoder-only architecture, and leverages multi-task learning to separate different layers into different functions of speech-to-text mapping, text-to-text generation, and text-to-speech mapping. During inference, Stream-Omni autoregressively generates speech at the outermost layer, while relying on the LLM backbone at the inner layers for response generation.
In this way, Stream-Omni preserves the generative capabilities and knowledge within the LLM core, while effectively broadening its interaction modalities, avoiding the high cost of using large-scale speech data to relearn textual knowledge. The speech interaction process in Stream-Omni includes speech tokenizer, speech-text mapping, text generation, and streaming speech generation.

\textbf{Speech Tokenizer}\quad To enable the mapping with text token, Stream-Omni employs the pre-trained CosyVoice speech tokenizer \citep{cosyvoice} to discretize the raw speech $S$ into a sequence of discrete speech units $U = (u_1, \cdots, u_{|U|})$:
\begin{gather}
    U = \mathrm{SpeechTokenizer}(S),
\end{gather}
where $\mathrm{SpeechTokenizer}(\cdot)$ denotes speech tokenizer, with the speech units vocabulary $\mathcal{V}^{\mathrm{U}}$. To joint modeling speech and text, we extend the vocabulary by merging the speech unit vocabulary $\mathcal{V}^{\mathrm{U}}$ with the LLM's text vocabulary $\mathcal{V}^{\mathrm{T}}$, and introduce a special blank token $\langle\text{blank}\rangle$, yielding the multimodal vocabulary of Stream-Omni $\mathcal{V}^{\mathrm{omni}} = \mathcal{V}^{\mathrm{T}} \cup \mathcal{V}^{\mathrm{U}} \cup \{\langle\text{blank}\rangle\}$.

\textbf{Speech-Text Mapping}\quad 
To take advantage of LLM's capabilities, Stream-Omni introduces the bottom and top speech layers to learn the speech-text mapping, thereby transferring the text capabilities within LLM to the speech modality. 
Specifically, the bottom and top speech layers consist of $N^{bottom}_{speech}$ and $N^{top}_{speech}$ Transformer layers, which share the same configuration as the LLM backbone. The bottom speech layers $\mathcal{F}^{bottom}_{speech}(\cdot)$ maps the speech units $U$ to the text:
\begin{gather}
    H^{\mathrm{U}} = \mathcal{F}^{bottom}_{speech}(U),
\end{gather}
where $H^{\mathrm{U}}$ denotes the representation of the speech units. Then, to achieve speech-to-text mapping, Stream-Omni introduces a Connectionist Temporal Classification (CTC) \citep{ctc} decoder $\mathrm{CTCDec}(\cdot)$ to decode the text sequence from $H^{\mathrm{U}}$:
\begin{gather}
    D^{\mathrm{U}} = \mathrm{CTCDec}(H^{\mathrm{U}}), \label{eq:ctcdec}
\end{gather}
where $D^{\mathrm{U}} \in \mathbb{R}^{|U| \times |\mathcal{V}^{\mathrm{omni}}|}$ represents the probability distribution over the multimodal vocabulary for each speech unit, which can be decoded into a CTC sequence that includes repeated and blank tokens. During training, this module is optimized using the CTC loss:
\begin{gather}
    \mathcal{L}_{CTC} = -\log \sum_{Z \in \Pi^{-1}(X)} p(Z \mid D^{\mathrm{U}}), \label{eq:4_ctc}
\end{gather}
where $\Pi^{-1}(X)$ denotes the set of all possible CTC sequences that map to the text sequence $X$ by removing repeated and blank tokens, and $p(Z \mid D^{\mathrm{U}})$ is the decoding probability of sequence $Z$ from $D^{\mathrm{U}}$. At inference time, Stream-Omni can decode the CTC sequence from $D^{\mathrm{U}}$ to produce streaming speech recognition results as an intermediate output for user. More potentially, the CTC decoder holds promise for real-time speech interaction by detecting when the user has stopped speaking based on the consecutive blank tokens in the CTC sequence \citep{zhang-etal-2024-streamspeech}.

\textbf{Text Generation}\quad Through CTC modeling, the bottom speech layers map the speech units into the text representation, achieving speech-text alignment at the representational level. To further bridge the structural gap between speech and text, Stream-Omni removes blank tokens $\langle\text{blank}\rangle$ from $H^{\mathrm{U}}$ to produce the refined sequence $\hat{H}^{\mathrm{U}}$. To preserve the model's understanding of the speech inputs, this blank token removal is only performed during the generation phase (i.e., generated speech).

The processed speech representation $\hat{H}^{\mathrm{U}}$ is then concatenated with the visual representation $H^{\mathrm{V}}$ (if has visual inputs) and fed into the LLM backbone $\mathcal{F}_{llm}(\cdot)$ to generate the text representation $H^{\mathrm{T}}$:
\begin{gather}
    H^{\mathrm{T}} = \mathcal{F}_{llm}([H^{\mathrm{V}} : \hat{H}^{\mathrm{U}}]),
\end{gather}
where $[\cdot : \cdot]$ is sequence concatenation. Owing to the semantic alignment via CTC modeling, Stream-Omni can transfer text intelligence to the speech modality while preserving the text capabilities. 

\textbf{Streaming Speech Generation}\quad While autoregressively generating the text outputs, Stream-Omni uses top speech layers to generate the corresponding speech units in a streaming manner. To ensure consistency between the generated speech and text, we introduce an alignment-based fusion to use text information to guide speech unit generation.

\begin{wrapfigure}{r}{0.385\textwidth} 
\vspace{-4mm}
    \centering
    \includegraphics[width=\linewidth]{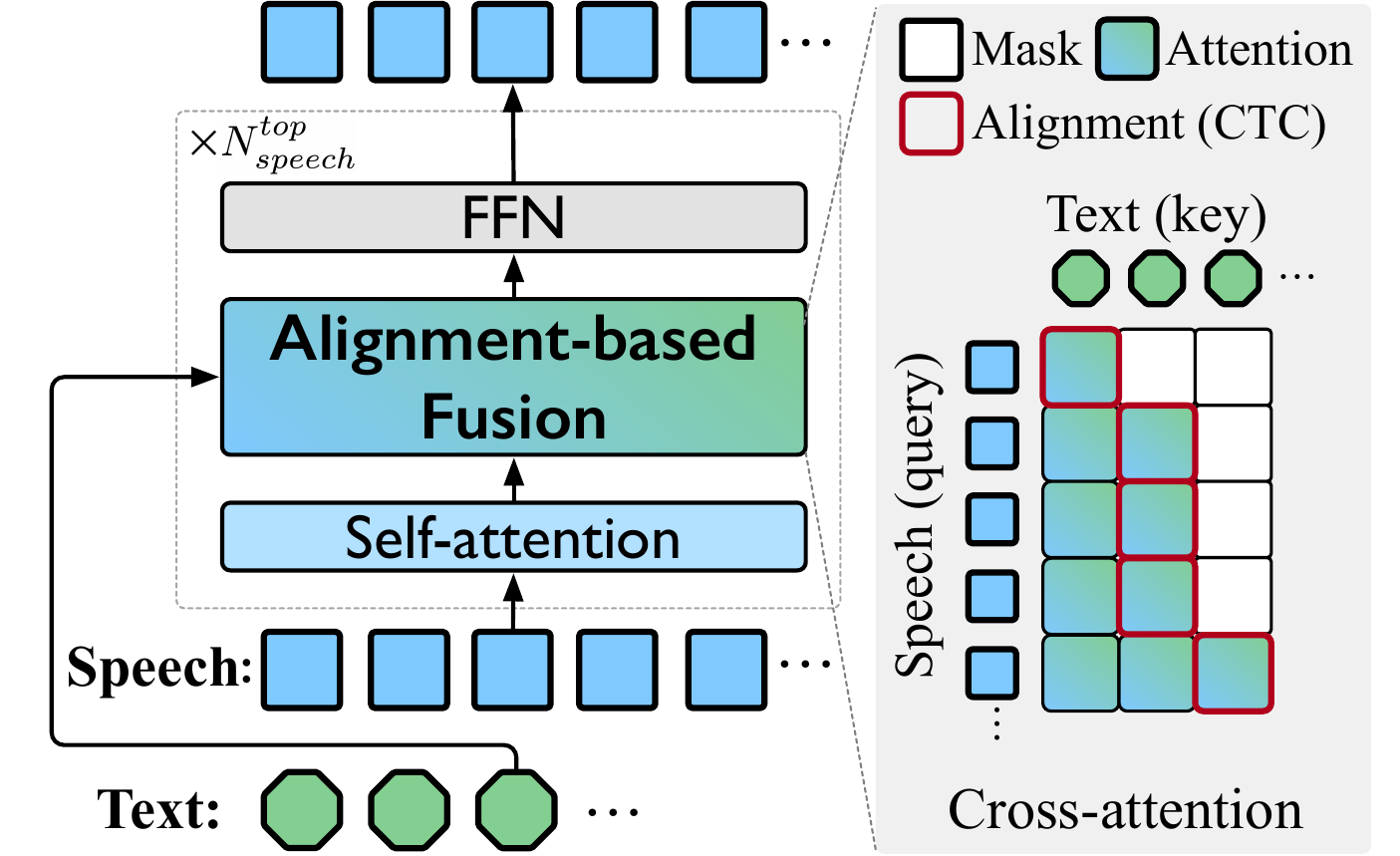}
    \vspace{-4mm}
    \caption{Diagram of top speech layers.}
    \label{fig:fusion}
    \vspace{-5mm}
\end{wrapfigure}
As illustrated in Figure~\ref{fig:fusion}, the top speech layers take the speech representations $H^{\mathrm{U}}$ from bottom speech layers and text representations $H^{\mathrm{T}}$ from the LLM backbone as the inputs, where each layer comprises self-attention, alignment-based fusion, and FFN. The alignment-based fusion module fuses the text representations $H^{\mathrm{T}}$ into the speech representations $H^{\mathrm{U}}$, thereby achieving text-to-speech mapping. However, to enable streaming generation, the key challenge lies in accurately identifying which text corresponds to each speech unit, thereby generating the speech units once the related text token is generated.

Fortunately, the CTC decoder introduced in Stream-Omni can naturally capture the positional alignment between speech and text \citep{zhang-etal-2024-streamspeech}, which can be used to guide the alignment-based fusion. Formally, based on the CTC sequence $D^{\mathrm{U}}$, Stream-Omni computes the number of aligned text tokens (excluding duplicate and blank tokens) corresponding to the speech sequence up to unit $u_i$, denoted as $\mathcal{N}_i$. That is, within the first $i$ speech units $U_{\leq i}$, Stream-Omni identifies the first $\mathcal{N}_i$ text tokens $X_{\leq \mathcal{N}_i}$. Accordingly, when autoregressively generating the next speech unit $u_{i+1}$, Stream-Omni should use the next text token $x_{\mathcal{N}_{i}+1}$ to guide the generation of speech unit $u_{i+1}$.

In practice, to involve richer text context, Stream-Omni extends the fusion window from the aligned text token $x_{\mathcal{N}_{i}+1}$ to its preceding $W - 1$ tokens, where $W$ is the hyperparameter of window size. The alignment-based fusion is implemented via cross-attention \citep{luo2025openomniadvancingopensourceomnimodal}, with the speech representations attending to the text representations, so the fused representation $h^{fusion}_i$ of speech unit $u_i$ is:
\begin{gather}
    h^{fusion}_{i} = \textrm{CrossAttn}\left(u_{i},\;H_{\mathcal{N}_{i}+2-W\;:\;\mathcal{N}_{i}+1}^{\mathrm{T}}\right), \label{eq:fusion}
\end{gather}
where $H^{\mathrm{T}}_{\mathcal{N}_i + 2 - W : \mathcal{N}_i + 1}$ are $W$ text representations within the local window ($W\!=\!5$ in Stream-Omni). To reduce generation latency, similar to the widely used wait-k policy in simultaneous translation \citep{ma-etal-2019-stacl,future-guided,zhang-feng-2021-universal,zhang-feng-2022-information,zhang-feng-2023-end}, Stream-Omni begins streaming speech generation after lagging $K$ text tokens ($K\!=\!3$ in Stream-Omni). Therefore, the first speech unit will be generated immediately after $K$ text tokens have been produced. Using the top speech layers $\mathcal{F}^{top}_{speech}(\cdot)$, Stream-Omni can simultaneously generate both text and the corresponding speech units:
\begin{gather}
    \hat{U} = \mathcal{F}^{top}_{speech}\left(H^{\mathrm{U}}, H^{\mathrm{T}}\right),
\end{gather}
where $\hat{U}$ denotes the generated speech unit sequence. Finally, a CosyVoice speech decoder \citep{cosyvoice} is used to synthesize the speech waveform from the generated speech units.

\subsection{Training}

Stream-Omni achieves efficient alignment across text, visual, and speech modalities, thus requiring only a small amount of tri-modal training data. Given the scarcity of existing datasets that jointly incorporate all three modalities, we first construct a tri-modal corpus consisting of text, images, and speech through an automated pipeline. Then, Stream-Omni adopts a three-stage training strategy to progressively align the text, visual, and speech modalities.

\subsubsection{Data Construction}

\begin{table}[t]
\caption{Training stages and data of Stream-Omni.}
  \label{tab:data}
\centering\footnotesize
\begin{tabular}{clll} \toprule
\textbf{Stages}                                                                                & \textbf{Training Tasks}                                                                                               & \textbf{Trainable Modules}                                                                        & \textbf{Datasets}                                                                         \\ \midrule
\multirow{3}{*}{\textbf{\begin{tabular}[c]{@{}c@{}}Stage1:\\ Vision-Text\end{tabular}}}        & \multirow{3}{*}{Vision$+$Text$\rightarrow$Text}                                                                                     & \multirow{3}{*}{\begin{tabular}[c]{@{}l@{}}Projection\\ LLM Backbone\end{tabular}}                & LLaVA                                                                                     \\
                                                                                               &                                                                                                                       &                                                                                                   & LLaVA-OV                                                                                  \\
                                                                                               &                                                                                                                       &                                                                                                   & LLaVA-zh                                                                                  \\\midrule
\multirow{6}{*}{\textbf{\begin{tabular}[c]{@{}c@{}}Stage2:\\ Speech-Text\end{tabular}}}        & \multirow{6}{*}{\begin{tabular}[c]{@{}l@{}}ASR (CTC Loss in Eq.(\ref{eq:fusion}))\\ Speech$\rightarrow$Speech\end{tabular}}                               & \multirow{6}{*}{\begin{tabular}[c]{@{}l@{}}Bottom Speech Layers\\ Top Speech Layers\end{tabular}} & LibriSpeech (960h)                                                                        \\
                                                                                               &                                                                                                                       &                                                                                                   & WenetSpeech (1240h)                                                                       \\
                                                                                               &                                                                                                                       &                                                                                                   & UltraChat$^{\text{tts}}$ (6500h)                                                                         \\
                                                                                               &                                                                                                                       &                                                                                                   & Wiki$^{\text{tts}}$ (4000h)                                                                              \\
                                                                                               &                                                                                                                       &                                                                                                   & LLaVA$^{\text{tts}}$ (8700h)                                                                             \\
                                                                                               &                                                                                                                       &                                                                                                   & LLaVA-zh$^{\text{tts}}$ (1200h)                                                                          \\\midrule
\multirow{3}{*}{\textbf{\begin{tabular}[c]{@{}c@{}}Stage3:\\ Text-Vision-Speech\end{tabular}}} & \multirow{3}{*}{\begin{tabular}[c]{@{}l@{}}Vision$+$Text$\rightarrow$Text\\ Vision$+$Speech$\rightarrow$Text\\ Vision$+$Speech$\rightarrow$Speech\end{tabular}} & \multirow{3}{*}{LLM Backbone}                                                                     & \multirow{3}{*}{\begin{tabular}[c]{@{}l@{}}LLaVA$^{\text{tts}}$ (8700h)\\ LLaVA-zh$^{\text{tts}}$ (1200h)\end{tabular}} \\
                                                                                               &                                                                                                                       &                                                                                                   &                                                                                           \\
                                                                                               &                                                                                                                       &                                                                                                   &                                                                                          \\\bottomrule   
\end{tabular}

\end{table}

The training of Stream-Omni involves text-vision, text-speech, and text-vision-speech multimodal datasets to support interactions across various modality combinations. For text-vision data, we adopt the LLaVA \citep{llava} and the LLaVA-OV dataset \citep{li2024llavaonevisioneasyvisualtask}, while filtering out samples involving maths, code, and other content unsuitable for speech interaction. For text-speech data, we use automatic speech recognition (ASR) corpora from LibriSpeech \citep{7178964} and WenetSpeech \citep{wenetspeech} to train bottom speech layers. Given the scarcity of public speech interaction data, we construct speech interaction dataset by converting existing text-only and vision-language instruction datasets into speech interactions datasets using open-source text-to-speech synthesis (TTS) \citep{cosyvoice}, named \emph{InstructOmni}\footnote{\url{https://huggingface.co/datasets/ICTNLP/InstructOmni}}. The construction details are introduced in Appendix \ref{app:train_data}. Table~\ref{tab:data} summarizes the used training data (only 23K hours of speech), where those marked with superscript `tts' indicate synthesized speech interaction dataset.

\subsubsection{3-Stage Training}

Stream-Omni is initialized using a LLM and adopts a three-stage training strategy, which aligns vision and speech with the text in succession, and then models alignments across three modalities.

\textbf{Stage 1: Vision-Text Alignment}\quad In this stage, Stream-Omni uses the standard training method used in vision-oriented LMMs such as LLaVA~\citep{llava}.

\textbf{Stage 2: Speech-Text Alignment}\quad In this stage, the speech-text alignment is achieved by training the bottom and top speech layers using a combination of CTC loss after the bottom speech layers (refer to Eq.(~\ref{eq:4_ctc})) and cross-entropy loss after the top speech layers. Note that, the text representations fed into the top speech layers during training (i.e., $H^{\mathrm{T}}$ in Eq.(\ref{eq:fusion})) are drawn from ground-truth transcriptions rather than LLM generated text, which aim to avoid text-speech dismatching \citep{luo2025openomniadvancingopensourceomnimodal} caused by generating incorrect text, thereby enhancing the consistency of text-to-speech generation.

\textbf{Stage 3: Text-Vision-Speech Alignment}\quad Finally, we train the LLM backbone of Stream-Omni using constructed tri-modal data through multi-task learning. Specifically, we formulate multiple tasks by combining different modalities, including Vision$+$Text$\rightarrow$Text, Vision$+$Speech$\rightarrow$Text, and Vision$+$Speech$\rightarrow$Speech, which are all optimized using the cross-entropy loss. In this way, Stream-Omni is able to flexibly support interactions under various modality combinations.

\subsection{Inference}

\begin{algorithm}[t]
\footnotesize
\caption{Inference of Stream-Omni}\label{Omnialgorithm}
\begin{algorithmic}[1]
\renewcommand{\algorithmicrequire}{\textbf{Input:}}
\Require Speech input $S$, Vision input $V$, Fusion window size $W$, Lagging text tokens $K$
\renewcommand{\algorithmicrequire}{\textbf{Output:}}
\Require Generated speech output $\widehat{S}$
\renewcommand{\algorithmicrequire}{\textbf{Init:}}
\Require ASR results (CTC sequence) $\widehat{A} = [~]$; Generated text tokens $\widehat{Y} = [~]$; Generated speech units $\widehat{U} = [~]$

\State Extract visual representation $H^{\mathrm{V}}$ from $V$ using the vision encoder and projection;
\State Extract speech units $U$ from $S$ using the speech tokenizer;
\State $H^{\mathrm{U}} \leftarrow \mathcal{F}^{bottom}_{speech}(U)$;\textcolor{red}{\Comment{\texttt{simultaneously produce ASR results of speech inputs}}}

\While{$\widehat{Y}[-1] \neq \left<eos\right>$}
    \State $y \leftarrow \mathcal{F}_{llm}([H^{\mathrm{V}} : \hat{H}^{\mathrm{U}} : \widehat{Y}])$
    \State $\widehat{Y}.\mathrm{append}(y)$; \textcolor{red}{\Comment{\texttt{simultaneously produce text outputs}}}

    \State \textbf{if} $|\widehat{Y}|<K$ \textbf{then} \textbf{continue}; \textcolor{gray}{\Comment{\texttt{lagging $K$ text tokens}}}

    \State \textcolor{gray}{// Generate speech units corresponding to $y$ until the text token is recognized in the generated speech}
    \While{$\widehat{A}[-1] == \left<blank\right>$ \textbf{or} $\widehat{A}[-1] == \widehat{A}[-2]$} \textcolor{blue}{\Comment{\texttt{generate speech for text $y$}}}
        \State Generate speech unit $u$ based on $H^{\mathrm{U}}$ and $\widehat{Y}[-W:]$ based on Eq.(\ref{eq:fusion});
        \State $\widehat{U}.\mathrm{append}(u)$;
        \State $a \leftarrow \mathrm{argmax}\left(\mathrm{CTCDec}(\mathcal{F}^{\mathrm{bottom}}_{\mathrm{speech}}(U))\right)$; \textcolor{gray}{\Comment{\texttt{recognize text from generated speech}}}
        \State $\widehat{A}.\mathrm{append}(a)$;
    \EndWhile

    \State Synthesize speech $s$ from $\widehat{U}$ using the speech decoder;
    \State $\widehat{S}.\mathrm{append}(s)$;
\EndWhile

\State \textbf{return} $\widehat{S}$
\end{algorithmic}
\end{algorithm}

Algorithm \ref{Omnialgorithm} gives the inference process of Stream-Omni when performing vision-grounded speech interaction. Given vision input $V$ and speech input $S$, Stream-Omni generates the text token $y$ in an autoregressive manner, and simultaneously synthesizes the corresponding speech of $y$. During speech synthesis, Stream-Omni autoregressively generates speech units $u$ based on $y$, until the entire speech corresponding to $y$ is generated. 
To determine whether the generated speech units for $y$ are complete, Stream-Omni leverages alignment in the CTC decoder (in Eq.(\ref{eq:ctcdec})). If the CTC decoder identifies a new text token from the generated $u$ (i.e., the semantics of the generated speech are complete), the model proceeds to generate the next text token. Otherwise, the model continues to generate speech units for the current $y$. Stream-Omni repeats the above process until $\left<\text{eos}\right>$ is generated.

Besides vision-grounded speech interaction, Stream-Omni also supports interaction of various modality combinations. As shown in Figure~\ref{fig:model}(right), by flexibly integrating the vision encoder, bottom speech layers, LLM, and top speech layers, Stream-Omni can support various multimodal scenarios.

\section{Experiments}
\label{sec:exp}
\subsection{Benchmarks}

We evaluate the multimodal capabilities of Stream-Omni across vision and speech benchmarks. For vision evaluation, we conduct experiments on 11 benchmarks used by LLaVA, including VQA-v2 ($\text{VQA}^{\text{v2}}$) \citep{Goyal_2017_CVPR}, GQA \citep{Hudson_2019_CVPR}, VizWiz \citep{Gurari_2018_CVPR}, ScienceQA-IMG (SciQA) \citep{NEURIPS2022_11332b6b}, TextVQA ($\text{VQA}^{\text{T}}$) \citep{Singh_2019_CVPR}, POPE \citep{li2023evaluating}, MME \citep{fu2024mmecomprehensiveevaluationbenchmark}, MMBench (MMB) \citep{liu2024mmbenchmultimodalmodelallaround}, SEED-Bench (SEED) \citep{Li_2024_CVPR}, LLaVA-Bench-in-the-Wild ($\text{LLaVA}^{\text{W}}$) \citep{NEURIPS2023_6dcf277e}, and MM-Vet \citep{yu2023mmvetevaluatinglargemultimodal}. All evaluations follow LLaVA~\citep{llava} to ensure comparability. For speech evaluation, we assess the model's knowledge-grounded speech interaction on spoken question answering benchmarks, Llama Questions (Llama Q.) \citep{nachmani2024spoken} and Web Questions (Web Q.) \citep{berant-etal-2013-semantic}, where the metric is the accuracy that whether the model's response matches the ground-truth answer. 

To further assess Stream-Omni's vision-grounded speech interaction capabilities, we construct a real-world visual-speech interaction benchmark based on the real-world VQA benchmark VisIT~\citep{visit}, named \emph{SpokenVisIT}\footnote{\url{https://huggingface.co/datasets/ICTNLP/SpokenVisIT}}. Following \citet{fang2025llamaomni}, the evaluation for SpokenVisIT employs the GPT model (gpt-4o version) to assign a score ranging from 1 to 5 for response. Appendix \ref{app:SpokenVisIT} gives the details of SpokenVisIT benchmark. Following previous works \citep{fang2025llamaomni,zeng2024glm4voiceintelligenthumanlikeendtoend}, all speech evaluations are further divided into speech-to-text (S$\rightarrow$T) and speech-to-speech (S$\rightarrow$S) settings. For generated speech responses, we use Whisper-large-v3\footnote{\url{https://huggingface.co/openai/whisper-large-v3}} \citep{radford2022robust} to transcribe the speech into text for evaluation.

\subsection{Baselines}

We compare Stream-Omni with vision-oriented, speech-oriented, and omni-modal LMMs of similar model scale and training data size. Vision-oriented LMM baselines include models comparable in scale to LLaVA-v1.5 \citep{llava}, such as BLIP-2 \citep{blip2}, InstructBLIP \citep{instructblip}, IDEFICS \citep{laurenccon2023introducing}, Qwen-VL \citep{bai2023qwenvlversatilevisionlanguagemodel}, Qwen-VL-Chat \citep{bai2023qwenvlversatilevisionlanguagemodel}, SPHINX \citep{lin2023sphinxjointmixingweights}, and mPLUG-Owl2 \citep{Ye_2024_CVPR}. Speech-oriented LMM baselines include TWIST \citep{hassid2023textually}, SpeechGPT \citep{zhang-etal-2023-speechgpt}, Spectron \citep{nachmani2024spoken}, Moshi \citep{défossez2024moshispeechtextfoundationmodel}, Freeze-Omni \citep{wang2024freezeomnismartlowlatency}, LLaMA-Omni \citep{fang2025llamaomni}, and GLM-4-Voice \citep{zeng2024glm4voiceintelligenthumanlikeendtoend}. Most existing omni-modal LMMs are trained on large-scale proprietary datasets \citep{li2024baichuanomnitechnicalreport,xu2025qwen25omnitechnicalreport,ji2025capybaraomniefficientparadigmbuilding}. For a fair comparison, we mainly compare Stream-Omni with VITA-1.5 \citep{fu2025vita15gpt4olevelrealtime}, a text-vision-speech LMM trained on a comparable amount of data, primarily based on LLaVA \citep{llava} and LLaVA-OV \citep{li2024llavaonevisioneasyvisualtask}. Additionally, we also compare Stream-Omni with some methods of similar data scale to demonstrate Stream-Omni's performance among advanced omni-modal LMMs, such AnyGPT \citep{zhan2024anygptunifiedmultimodalllm}, EMOVA \citep{chen2025emovaempoweringlanguagemodels} and OpenOmni \cite{luo2025openomniadvancingopensourceomnimodal}. Note that these models were trained using different datasets and training pipelines with Stream-Omni.

\begin{table}[t]
    \centering\scriptsize
    \caption{Results on visual understanding benchmarks.}
    \label{tab:vision_eval}
    \begin{tabular}{llC{0.4cm}C{0.4cm}C{0.4cm}C{0.4cm}C{0.4cm}C{0.4cm}C{0.5cm}C{0.4cm}C{0.4cm}C{0.4cm}C{0.4cm}C{0.4cm}}\toprule
\textbf{Methods}      & \textbf{LLM}                   & $\!\!$$\textbf{\text{VQA}}^{\!\text{v2}}$ & $\!\!$\textbf{GQA} & \textbf{\begin{tabular}[c]{@{}c@{}}Vis\\ Wiz\end{tabular}} & \textbf{\begin{tabular}[c]{@{}c@{}}Sci\\ QA\end{tabular}} & $\!\!$$\textbf{\text{VQA}}^{\!\text{T}}$ & $\!\!$\textbf{POPE} & \textbf{MME} & $\!\!$\textbf{MMB} & $\!\!$\textbf{SEED} & $\!\!\!$$\textbf{\text{LLaVA}}^{\text{\!\!W}}$ & \textbf{\begin{tabular}[c]{@{}c@{}}MM-\\ Vet\end{tabular}} & \textbf{\begin{tabular}[c]{@{}c@{}}Avg.\\ (\%)\end{tabular}} \\\midrule
\textbf{BLIP-2}       & Vicuna-13B            & 65.0  & 41.0 & 19.6   & 61.0  & 42.5    & 85.3 & 1293.8 & –    & 46.4 & 38.1  & 22.4   & –    \\
\textbf{InstructBLIP} & Vicuna-7B             & –     & 49.2 & 34.5   & 60.5  & 50.1    & –    & –      & 36.0 & 53.4 & 60.9  & 26.2   & –    \\
\textbf{IDEFICS-9B}   & LLaMA-7B              & 50.9  & 38.4 & 35.5   & –     & 25.9    & –    & –      & 48.2 & –    & –     & –      & –    \\
\textbf{Qwen-VL}      & Qwen-7B               & 78.8  & 59.3 & 35.2   & 67.1  & 63.8    & –    & –      & 38.2 & 56.3 & –     & –      & –    \\
\textbf{Qwen-VL-Chat} & Qwen-7B               & 78.2  & 57.5 & 38.9   & 68.2  & 61.5    & –    & 1487.5 & 60.6 & 58.2 & –     & –      & –    \\
\textbf{SPHINX}       & LLaMA-13B             & 78.1  & 62.6 & 39.9   & 69.3  & 51.6    & 80.7 & 1476.1 & 66.9 & 56.2 & 73.5  & 36.0   & 56.0 \\
\textbf{SPHINX-2k}    & LLaMA-13B             & 80.7  & 63.1 & 44.9   & 70.6  & 61.2    & 87.2 & 1470.6 & 65.9 & 57.9 & 76.9  & 40.2   & 59.0 \\
\textbf{mPLUG-Owl2}   & LLaMA-7B              & 79.4  & 56.1 & 54.5   & 68.7  & 54.3    & –    & 1450.2 & 64.5 & 57.8 & -     & 36.2   & –    \\
\textbf{LLaVA-1.5}    & Vicuna-7B             & 78.5  & 62.0 & 50.0   & 66.8  & 58.2    & 85.9 & 1510.7 & 64.3 & 58.6 & 63.4  & 30.5   & 56.3 \\
\textbf{LLaVA-NeXT} & Vicuna-7B & 81.8	&64.2	&57.6	&70.1	&64.9	&86.5	&1519.0	&67.4	&70.2	&81.6	&43.9	&62.6 \\
\textbf{LLaVA-OV} & Qwen2-7B & –	&–	&–	&96.0	&–	&–	&1580.0	&80.8	&75.4	&–	&–	&– \\\midrule
\textbf{EMOVA} & Qwen2.5-7B & –	&–	&–	&96.4	&–	&–	&–	&83.0	&75.5	&–	&59.4	&– \\
\textbf{OpenOmni} & Qwen2.5-7B & –	&–	&–	&–	&–	&–	&–	&76.2	&–	&–	&–	&– \\
\textbf{VITA-1.5}     & Qwen2-7B              & 78.8  & 60.6 & 54.8   & 90.9  & 65.0    & 85.7 & 1687.7 & 76.7 & 70.4 & 71.0  & 49.6   & 64.0 \\\midrule
\textbf{Stream-Omni}  & LLaMA-3.1-8B & 79.7  & 68.3 & 45.5   & 93.4  & 62.7    & 86.0 & 1752.7 & 82.4 & 76.3 & 71.2  & 44.7   & 64.7\\\bottomrule
\end{tabular}
\end{table}

\subsection{Configuration}

Stream-Omni is built upon the LLaMA-3.1-8B-Instruct\footnote{\url{https://huggingface.co/meta-llama/Llama-3.1-8B-Instruct}} \citep{llama3}, which consists of 32 Transformer layers. For vision, Stream-Omni employs the SigLIP-so400m-patch14-384\footnote{\url{https://huggingface.co/google/siglip-so400m-patch14-384}} \citep{Zhai_2023_ICCV} as the vision encoder. For speech, Stream-Omni incorporates the bottom speech layers with 3 Transformer layers and top speech layers with 5 Transformer layers, where all Transformer layers share the same architecture and parameter configuration as those in LLM. The speech tokenizer and flow-matching-based speech decoder are adopted from CosyVoice-300M-25Hz\footnote{\url{https://modelscope.cn/models/iic/CosyVoice-300M-25Hz}} \citep{cosyvoice}. The vocabulary of Stream-Omni comprises 128K text tokens from LLaMA-3.1-8B-Instruct, 4096 speech units from the CosyVoice tokenizer, and a blank token $\left<\text{blank}\right>$. Stream-Omni is trained using 8 H800 GPUs and tested on 1 A100 GPU.

\section{Results and Analyses}
\label{sec:res}

\subsection{Visual Understanding}

We evaluate the visual understanding capabilities of Stream-Omni in Table~\ref{tab:vision_eval}. Compared to advanced vision-oriented LMMs and VITA-1.5 \citep{fu2025vita15gpt4olevelrealtime}, Stream-Omni demonstrates strong visual capabilities on various visual tasks.
More importantly, despite being a unified model that simultaneously supports vision, speech, and text, Stream-Omni achieves performance comparable to vision-oriented LMMs, indicating its effectiveness in mitigating modality interference.

\subsection{Speech Interaction}

\setlength{\columnsep}{5pt}
\begin{wraptable}{r}{0.5\textwidth}
\vspace{-4mm}
\caption{Results on spokenQA benchmarks.}
\vspace{-1mm}
\label{tab:spokenqa}
\centering
\scriptsize
\begin{tabular}{lC{0.4cm}C{0.4cm}C{0.4cm}C{0.4cm}C{0.4cm}C{0.4cm}}\toprule
\multirow{2}{*}{\textbf{Methods}} & \multicolumn{2}{c}{\textbf{Llama Q.}} & \multicolumn{2}{c}{\textbf{Web Q.}} & \multicolumn{2}{c}{\textbf{Avg.}}       \\\cmidrule(lr){2-3}\cmidrule(lr){4-5}\cmidrule(lr){6-7}
                         & S$\rightarrow$T              & S$\rightarrow$S              & S$\rightarrow$T             & S$\rightarrow$S             & S$\rightarrow$T           & S$\rightarrow$S           \\\midrule
\textbf{TWIST}                    & -                & 4.0              & -               & 1.5             & -             & 2.8           \\
\textbf{SpeechGPT}                & 21.6             & -                & 6.5             & -               & 14.1          & -             \\
\textbf{Spectron}                 & 21.9             & -                & 6.1             & -               & 14.0          & -             \\
\textbf{Moshi}                    & 62.3             & 21.0             & 26.6            & 9.2             & 44.5          & 15.1          \\
\textbf{GLM-4-Voice}              & 64.7             & \underline{50.7}             & 32.2            & 15.9            & 48.5          & 33.3          \\
\textbf{Freeze-Omni}              & 72.0             & -                & \textbf{44.7}   & -               & 58.4          & -             \\
\textbf{LLaMA-Omni}               & 67.7             & 49.0             & 33.4            & \underline{23.7}   & 50.6          & \underline{36.4}          \\
\textbf{VITA-1.5}                 & \textbf{76.7}    & -                & 42.7            & -               & \underline{59.7}          & -             \\ \midrule
\textbf{Stream-Omni}              & \underline{76.3}             & \textbf{65.0}    & \underline{44.2}            & \textbf{27.5}            & \textbf{60.3} & \textbf{46.3}\\\bottomrule             
\end{tabular}
\vspace{-2mm}
\end{wraptable}

To verify whether Stream-Omni can acquire speech capabilities and knowledge with a small amount of speech data, we conduct experiments on knowledge-based LLaMA Question and Web Question, covering both speech-to-text (S$\rightarrow$T) and speech-to-speech (S$\rightarrow$S) tasks. As shown in Table~\ref{tab:spokenqa}, Stream-Omni demonstrates strong knowledge-based speech interaction performance. Speech-oriented LMMs based on discrete speech units, such as SpeechGPT, Moshi, and GLM-4-Voice, typically rely on speech pretraining to acquire knowledge from large-scale speech data \citep{zhang-etal-2023-speechgpt,défossez2024moshispeechtextfoundationmodel,zeng2024glm4voiceintelligenthumanlikeendtoend}, Stream-Omni achieves superior knowledge-based speech interaction with significantly less speech data of 23K hours, particularly in the speech-to-text setting. This advantage primarily stems from the CTC-based speech-to-text mapping in Stream-Omni, which effectively transfers the text knowledge within LLM to the speech modality and thereby supports knowledge-based speech interaction in more efficient manner.

\subsection{Vision-grounded Speech Interaction}
\label{sec:SpokenVisIT}

\setlength{\columnsep}{10pt}
\begin{wraptable}{r}{0.38\textwidth}
\vspace{-4mm}
\caption{Results on SpokenVisIT (`V': vision, `T': text, `S': speech).}
\vspace{-1mm}
\label{tab:SpokenVisIT}
\centering
\scriptsize
\begin{tabular}{lC{0.7cm}C{0.7cm}C{0.7cm}}\toprule
\multirow{2}{*}{\textbf{Methods}} & \multicolumn{3}{c}{\textbf{SpokenVisIT}} \\ \cmidrule(lr){2-4}
                                  & $\!\!\!$V$+$T$\rightarrow$T         & $\!\!$V$+$S$\rightarrow$T        & $\!$V$+$S$\rightarrow$S        \\\midrule
\textbf{GPT-4V}                   & 4.81         & -           & -           \\\midrule
\textbf{VITA-1.5}                 & 3.63         & 3.45        & -           \\
\textbf{Stream-Omni}              & 3.93          & 3.68         & 2.62       \\\bottomrule
\end{tabular}
\vspace{-2mm}
\end{wraptable}

Most existing benchmarks for evaluating the vision-grounded speech interaction typically use multiple-choice formats, which do not align well with real-world application scenarios. To address this, we constructed SpokenVisIT based on VisIT-Bench \citep{visit}, a vision-grounded speech interaction benchmark based on real-world scenarios. We evaluate Stream-Omni on the SpokenVisIT benchmark in Table~\ref{tab:SpokenVisIT}. As the omni-modal LMMs with similar training data, Stream-Omni demonstrates superior real-world visual understanding capabilities compared to VITA-1.5. In addition, Stream-Omni supports speech generation, extending its potential for multimodal interaction. Appendix \ref{app:case} gives specific case studies, demonstrating the advantages of Stream-Omni's speech-text mapping in cross-modal consistency.


\subsection{Quality of Speech-Text Mapping}

\setlength{\columnsep}{5pt}
\begin{wraptable}{r}{0.51\textwidth}
\vspace{-12mm}
\caption{Results on LibriSpeech benchmarks.}
\vspace{-1mm}
\label{tab:asr}
\centering
\scriptsize
\begin{tabular}{lC{0.3cm}C{0.3cm}cC{0.3cm}c}\toprule
\multirow{2}{*}{\textbf{Methods}} & \multirow{2}{*}{$\!\!\!\!\!\!$\textbf{\begin{tabular}[c]{@{}c@{}}Stream\\ -ing\end{tabular} }} & \multicolumn{2}{c}{\textbf{test-clean}}                              & \multicolumn{2}{c}{\textbf{test-other}}                              \\ 
                                  &                                     & $\!\!$WER  & \begin{tabular}[c]{@{}c@{}}Inference\\ Time (ms)\end{tabular} & $\!\!$WER  & \begin{tabular}[c]{@{}c@{}}Inference\\ Time (ms)\end{tabular} \\ \midrule
\textbf{Whisper}                  & ×                                   & 2.5  & 692                                                           & 4.5  & 616                                                           \\
\textbf{SpeechGPT}                & ×                                   & $\!\!$18.9 & 794                                                           & $\!$29.1 & 755                                                           \\
\textbf{Moshi}                    & \checkmark                                   & 5.7  & -                                                             & -    & -                                                             \\
\textbf{Mini-Omni}               & ×                                   & 4.7  & 196                                                           & 9.4  & 148                                                           \\
\textbf{Freeze-Omni}             & ×                                   & 3.2  & 984                                                           & 7.7  & 965                                                           \\
\textbf{GLM-4-Voice}              & ×                                   & 2.8  & 756                                                           & 7.7  & 701                                                           \\\midrule
\textbf{AnyGPT}              & ×                                   & 8.5  & -                                                           & -  & -                                                           \\
\textbf{EMOVA}              & ×                                   & 4.1  & -                                                           & -  & -                                                           \\
\textbf{OpenOmni}              & ×                                   & 3.1  & -                                                           & 4.1  & -                                                           \\
\textbf{VITA-1.5}                 & ×                                   & 3.4  & -                                                             & 7.5  & -                                                             \\ \midrule
\textbf{Stream-Omni}              &  \checkmark                                    & 3.0  & 125                                                            & 7.2  & 104        \\\bottomrule                                                   
\end{tabular}
\vspace{-4mm}
\end{wraptable}

Stream-Omni introduces the auxiliary ASR task to train the bottom speech layers and CTC decoder, thereby learning effective speech-to-text mapping. To evaluate the quality of mapping, we evaluate the ASR performance of Stream-Omni on the LibriSpeech benchmark \citep{7178964}. As shown in Table~\ref{tab:asr}, Stream-Omni achieves advantages in both accuracy and inference time. SpeechGPT \citep{zhang-etal-2023-speechgpt}, Freeze-Omni \citep{wang2024freezeomnismartlowlatency}, and GLM-4-Voice \citep{zeng2024glm4voiceintelligenthumanlikeendtoend} need to forward full LMM to autoregressively generating the ASR results. In contrast, Stream-Omni generates the ASR results using its bottom speech layers in a non-autoregressive manner, resulting in lower inference time for ASR task. More importantly, this layer-dimension allows Stream-Omni to simultaneously present intermediate ASR results during speech interaction, providing users with a more comprehensive interaction experience.

\subsection{Effect of Alignment-based Fusion}

\setlength{\columnsep}{6pt}
\begin{wraptable}{r}{0.5\textwidth}
\vspace{-4mm}
\caption{Analysis on alignment-based fusion.}
\vspace{-1mm}
\label{tab:ab_fusion}
\centering
\scriptsize
\begin{tabular}{lccc} \toprule
\textbf{\begin{tabular}[c]{@{}l@{}}Fusion\\ Type\end{tabular}} & \textbf{\begin{tabular}[c]{@{}c@{}}Fusion \\ Window\end{tabular}} & \textbf{\begin{tabular}[c]{@{}c@{}}Llama Q.\\ S$\rightarrow$S\end{tabular}} & \textbf{\begin{tabular}[c]{@{}c@{}}Web Q.\\ S$\rightarrow$S\end{tabular}} \\ \midrule
\textbf{Attention}                                                  & 5                                                                 & 65.0                                                            & 27.5                                                          \\\midrule
\textbf{Add (input)}                                           & 1                                                                 & 40.3                                                            & 19.2                                                          \\
\textbf{Add (per layer)}                                       & 1                                                                 & 45.3                                                            & 21.5                                                          \\\midrule
\textbf{Attention}                                                  & 2                                                                 & 54.3                                                            & 22.1                                                          \\
\textbf{Attention}                                                  & 10                                                                & 62.3                                                            & 25.7                                                          \\
\textbf{Attention}                                                  & $\infty$                                                               & 60.0                                                            & 24.3      \\\bottomrule                                                   
\end{tabular}
\vspace{-2mm}
\end{wraptable}

Stream-Omni generates speech from text in a streaming manner using alignment-based fusion. To evaluate its effectiveness, we conduct the ablation study of alignment-based fusion on Llama Questions and Web Questions benchmarks (S$\rightarrow$S) in Table \ref{tab:ab_fusion}, focusing on the fusion type and the fusion window.

\textbf{Fusion Type}\quad For the fusion type, we compare the current cross-attention (named ``Attention'') with adding aligned text representations to the input (named ``Add (input)'') or each layer (named ``Add (per layer)'')  of the top speech layers. Results show that the attention-based approach outperforms the others, mainly due to its ability to attend to a broader context rather than merely adding a single text token. Existing speech-oriented LMMs \citep{fang2025llamaomni,fang2025llamaomni2llmbasedrealtimespoken} or omni-modal LMMs \citep{luo2025openomniadvancingopensourceomnimodal} often mix speech and text representations at the input of the speech decoder to achieve text-to-speech generation. Different with existing methods, Stream-Omni integrates the corresponding textual information into the speech representations at each layer of the speech decoder through alignment-based fusion, enabling high-quality text-to-speech generation.

\textbf{Fusion Window}\quad For the fusion window, we find that attending to either very few or all text tokens during speech generation is less effective than focusing on a moderate window of tokens, which is attributed to the inherent monotonicity and locality in text-to-speech generation. This is also in line with the widely used speech-text interleaved generation methods \citep{nguyen-etal-2025-spirit,zeng2024glm4voiceintelligenthumanlikeendtoend,zeng2024scalingspeechtextpretrainingsynthetic}. The difference lies in that previous methods achieve consistency between generated speech and the current text through interleaving along the sequence dimension, while alignment-based fusion ensures consistency by guiding the speech to attend to the current text along the layer dimension.

\section{Conclusion}

We propose Stream-Omni, a LMM that simultaneously supports various multimodal interactions. 
Stream-Omni achieves efficient modality alignments via the sequence-dimension concatenation for vision and layer-dimension mapping for speech. Furthermore, Stream-Omni can enhance the multimodal experience by simultaneously providing intermediate text results during speech interaction.

\section*{Limitations}
\label{app:limitation}

In this paper, we present Stream-Omni, a large multimodal model that supports text, vision, and speech. To address the scarcity of public tri-modal data, we focus on how to model the modality alignment more purposely to achieve efficient and flexible modality alignments. However, beyond the modeling way of modality alignments, high-quality multimodal interaction also rely on other factors, such as speech expressiveness and the degree of human-likeness. These aspects are important but are not the primary focus of Stream-Omni, so we leave them for future work.

\bibliographystyle{unsrtnat}
\bibliography{neurips_2025}

\newpage
\appendix

\section{Construction of InstructOmni}
\label{app:train_data}

Existing publicly available text and vision instruction data are readily accessible, while speech instruction data and tri-modal instruction data involving text, vision, and speech remain relatively scarce. To address this, we propose InstructOmni, an omni-modal dataset automatically constructed using text-to-speech (TTS) synthesis. InstructOmni builds upon existing publicly available text-only and vision-language instruction datasets by generating corresponding speech based on textual instructions and responses, thereby producing both speech-based instruction data and tri-modal instruction data for training.
Specifically, we synthesize speech instruction data from the LLaVA visual instruction tuning dataset \citep{llava}, the UltraChat text instruction tuning dataset \citep{ultrachat} (used in LLaMA-Omni \citep{fang2025llamaomni} and LLaMA-Omni2 \citep{fang2025llamaomni2llmbasedrealtimespoken}), and a subset of Wikipedia entries. The text instructions and responses from these sources are converted into speech using the CosyVoice TTS model \citep{cosyvoice}. To better simulate the variability of speech input in real-world scenarios, we randomly sample speaker embeddings from LibriSpeech \citep{7178964} and AISHELL \citep{AISHELL-3_2020}, and apply voice cloning techniques to generate speech with diverse speaker characteristics, thereby enhancing the realism and diversity of the speech.

Table~\ref{tab:data} summarizes the used datasets during training, where those marked with a superscript `tts' indicate samples with synthesized speech. Overall, Stream-Omni is trained on only 23K hours of speech data, which is significantly less than the large-scale datasets used in previous methods, such as TWIST (150K hours) \citep{hassid2023textually}, SpeechGPT (60K hours) \citep{zhang-etal-2023-speechgpt}, Moshi (7M hours) \citep{défossez2024moshispeechtextfoundationmodel}, GLM-4-Voice (700K hours) \citep{zeng2024glm4voiceintelligenthumanlikeendtoend}, and VITA-1.5 (110K hours) \citep{fu2025vita15gpt4olevelrealtime}, highlighting its advantage in data efficiency.

\section{Construction of SpokenVisIT}
\label{app:SpokenVisIT}

In Sec.\ref{sec:SpokenVisIT}, to align with real-world application scenarios, we construct SpokenVisIT benchmark based on VisIT-Bench \citep{visit} to evaluate the vision-grounded speech interaction capability of omni-modal LMMs. Here, we give a detailed introduction to SpokenVisIT.

To better reflect real-world scenarios of vision-based speech interaction, we adopt the VisIT-Bench~\citep{visit} as the source dataset (an open-ended generation format instead of multi-choice format is much suitable for real-world scenarios). VisIT is a real-world visual question answering benchmark comprising 574 images and 70 types of instructions covering object recognition, visual reasoning, creative writing, and more. Unlike existing vision evaluation benchmarks that mainly use multiple-choice format, all text instructions in the VisIT benchmark are written in a colloquial style, making it particularly well-suited for speech interaction. To adapt VisIT for speech interaction, we employ text-to-speech synthesis \citep{cosyvoice} to convert each text instruction into a corresponding speech utterance, resulting in a derived benchmark named SpokenVisIT. During construction, eight math-related instructions that were unsuitable for speech interaction were removed. For the evaluation metric, following the open-ended spoken interaction evaluation protocol proposed by~\citet{fang2025llamaomni}, we use ChatGPT (gpt-4o version) to assess the quality of responses on a 1-5 scale. The evaluation prompt includes the image caption as a reference, along with the question and the model's answer.

\begin{tcolorbox}
[title=Prompt of SpokenVisIT Evaluation ,colback=blue!10,colframe=blue!50!black,arc=1mm,boxrule=1pt,left=1mm,right=1mm,top=1mm,bottom=1mm, fonttitle=\scriptsize]
\footnotesize
I need your assistance in evaluating the performance of several models in a vision-based speech interaction scenario. These models process the user's spoken input and generate spoken responses. For evaluation purposes, both the user's speech input and the model's speech output have been transcribed into text using Automatic Speech Recognition (ASR). Additionally, a brief image caption is provided to help you understand the visual context of the conversation. Your task is to assess the model's responses based on the given visual context [Image Caption], the transcribed user input [Instruction], and the transcribed model output [Response]. Please evaluate the responses considering factors such as helpfulness, responsiveness, empathy, and suitability for real-world multimodal interaction, and assign a single score on a scale from 1 to 5.\\

Below are the image caption and the transcriptions of the user's instruction and the model's response:

\#\#\# [Image Caption]: \{caption\}

\#\#\# [Instruction]: \{question\}

\#\#\# [Response]: \{answer\}\\

After evaluating, please output the scores in JSON format: \{score: ...\}. You don't need to provide any explanations. 
\end{tcolorbox}

\section{Case Study}
\label{app:case}
\begin{figure}[t]
    \centering
    \includegraphics[width=\linewidth]{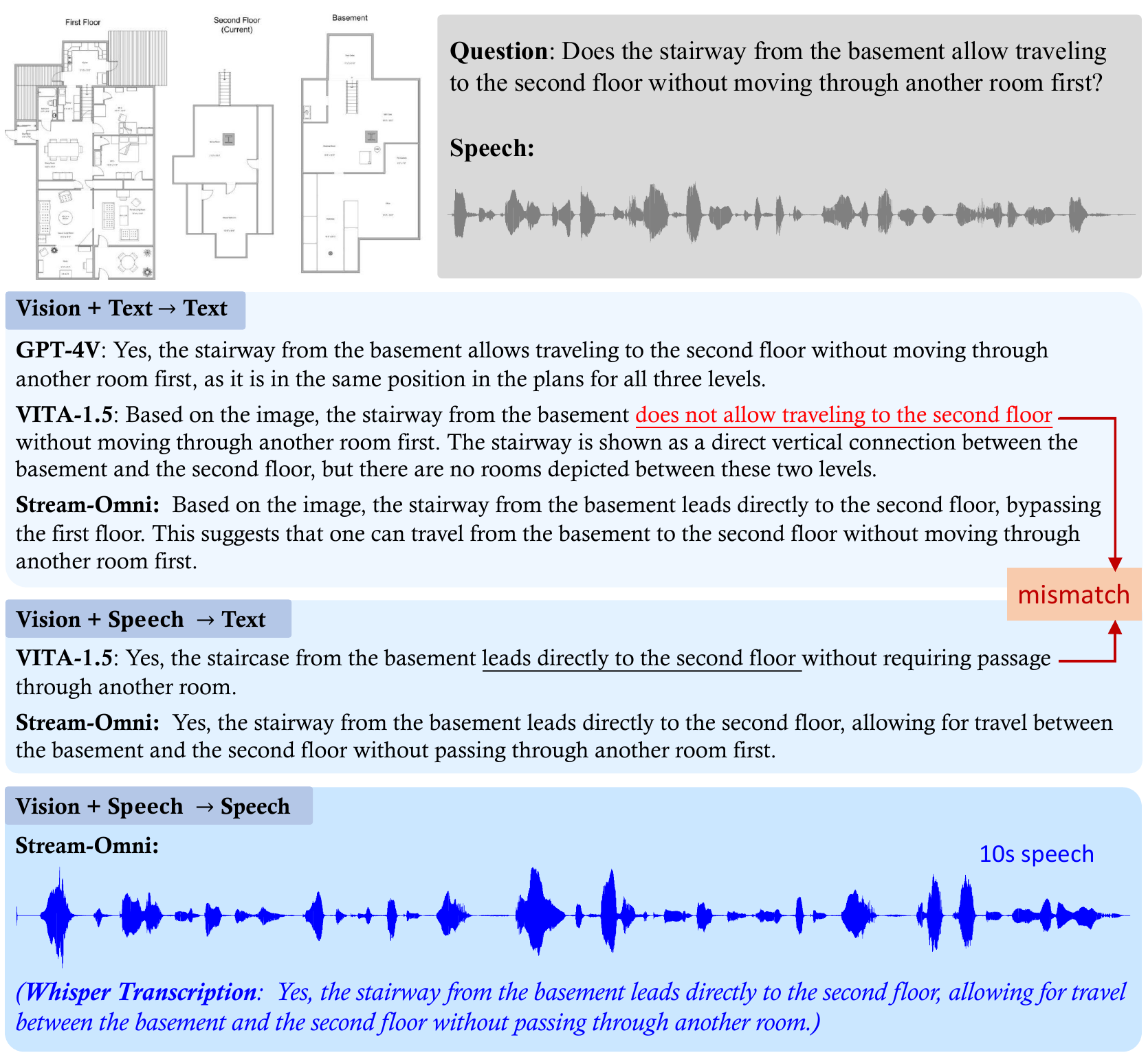}
    \caption{Case Study of Stream-Omni (detail understanding). }
    \label{fig:case1}
\end{figure}

\begin{figure}[t]
    \centering
    \includegraphics[width=\linewidth]{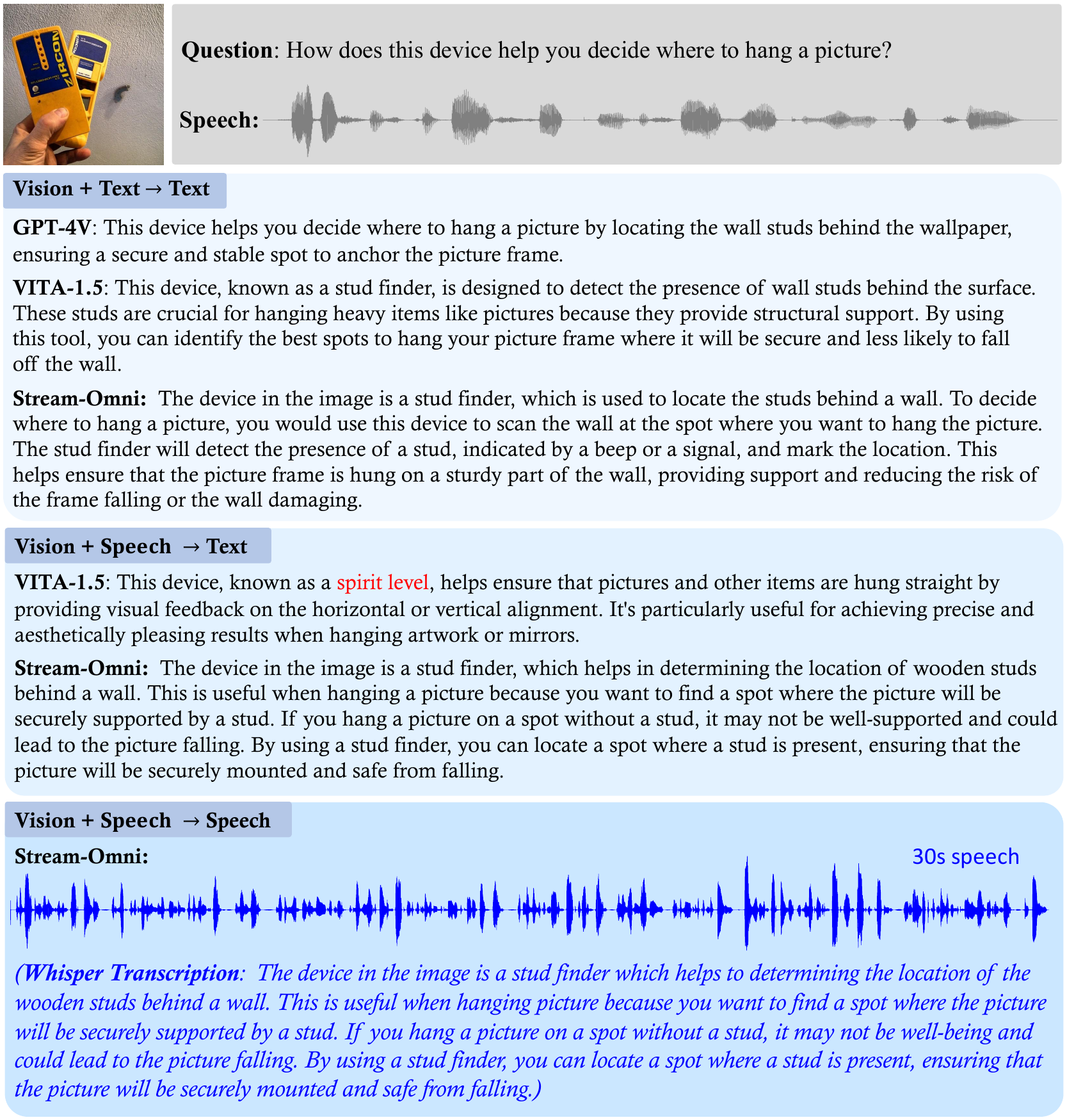}
    \caption{Case Study of Stream-Omni (long response). }
    \label{fig:case2}
\end{figure}

To provide a more intuitive demonstration of Stream-Omni's multimodal interaction capabilities, we conduct two case studies in Figure~\ref{fig:case1} and~\ref{fig:case2}, where both the visual and speech inputs are sourced from the constructed SpokenVisIT benchmark. The case in Figure~\ref{fig:case1} focuses on visual detail understanding, while the case in Figure~\ref{fig:case2} highlights the model's ability to generate long speech responses. The red-marked text indicates the incorrect part of the response. In both cases, Stream-Omni demonstrates good performance across different modalities. Specifically, in vision-based text interaction, Stream-Omni accurately interprets visual inputs and generates output sequences that closely resemble those produced by GPT-4V~\citep{gpt-4v}. When conditioned on both visual and speech inputs, Stream-Omni outperforms VITA-1.5.

In the example shown in Figure~\ref{fig:case1}, when the instruction is delivered via text and speech respectively, VITA-1.5 produces two contradictory responses of "\textit{does not allow traveling to the second floor}" and "\textit{leads directly to the second floor}". This contradictory response when facing different modal instructions stems from VITA-1.5's sequence-dimension concatenation of visual, speech, and text representations to achieve multimodal alignment~\citep{fu2025vita15gpt4olevelrealtime}, without modeling rigorous semantic alignment between the speech and text modalities. In contrast, Stream-Omni employs the speech-to-text mapping that enables precise semantic alignment between speech and text representations. As a result, Stream-Omni achieves more consistent performance across modalities and can generate similar responses regardless of whether the instruction is delivered via text or speech.

In the example shown in Figure~\ref{fig:case2}, Stream-Omni exhibits strong speech generation capabilities, producing high-quality speech outputs lasting up to 30 seconds. Notably, the generated speech is highly consistent with the corresponding text outputs, underscoring the effectiveness of the proposed alignment-based fusion module. Overall, Stream-Omni enables high-quality, vision-grounded speech interactions, fulfilling the diverse requirements of multimodal interaction.



\end{document}